\documentclass{article}

\usepackage{microtype}
\usepackage{graphicx}
\usepackage{booktabs} %

\usepackage[pagebackref,breaklinks,colorlinks]{hyperref}

\usepackage[dvipsnames]{xcolor}
\usepackage[accepted]{icml2023}

\usepackage{amsmath}
\usepackage{amssymb}
\usepackage{mathtools}
\usepackage{amsthm}
\usepackage{bm}
\usepackage{enumitem}
\usepackage{mathrsfs}

\usepackage[textwidth=18mm]{todonotes}
\usepackage[capitalize]{cleveref}

\usepackage{xspace}
\usepackage{colortbl}
\usepackage{subcaption}
\usepackage{bbm}

\crefname{section}{Sec.}{Secs.}
\Crefname{section}{Section}{Sections}
\Crefname{table}{Table}{Tables}
\crefname{table}{Tab.}{Tabs.}
\DeclareMathOperator{\diversity}{Diversity}

\definecolor{Gray}{gray}{0.5}
\newcolumntype{a}{>{\color{Gray}}c}
\definecolor{NewBlue}{rgb}{0.95, 0.95, 1.0}
\newcolumntype{b}{>{\columncolor{NewBlue}}c}

\icmltitlerunning{Adaptive Coordination for Social Embodied Rearrangement}

\begin{document}
\DeclarePairedDelimiter\abs{\lvert}{\rvert}
\newcommand{\cagent}{coordination agent\xspace}
\newcommand{\Cagent}{Coordination agent\xspace}
\newcommand{\cag}{coordination agent\xspace}
\newcommand{\tname}{Social Rearrangement\xspace}
\newcommand{\Tname}{Social Rearrangement\xspace}
\newcommand{\HcUnseen}{Scripted Unseen\xspace}
\newcommand{\NnUnseen}{Learned Unseen\xspace}

\newcommand{\oracle}{oracle\xspace}
\newcommand{\Oracle}{Oracle\xspace}
\newcommand{\CGain}{Cooperation Efficiency Gain\xspace}
\newcommand{\cgain}{cooperation efficiency gain\xspace}
\newcommand{\eval}{ZSC eval\xspace}
\newcommand{\Eval}{ZSC eval\xspace}
\newcommand{\train}{train-pop eval\xspace}
\newcommand{\Train}{Train-pop eval\xspace}
\newcommand{\traintitle}{Train-Pop Eval\xspace}
\newcommand{\evaltitle}{ZSC Eval\xspace}

\newcommand\asz[1]{\textcolor[rgb]{0.2, 0.6, 0.86}{AS: #1}}
\newcommand{\asn}[1]{\todo[color=red!20, size=\tiny]{AS: #1}}
\newcommand{\aks}[1]{\textcolor{orange}{Akshara: #1}}
\newcommand{\rd}[1]{\textcolor{blue}{RD: #1}}
\newcommand{\RM}[1]{\todo[color=blue!20, size=\tiny]{RM: #1}}
\newcommand{\zk}[1]{{\color{blue}{(\textbf{ZK: }#1)}}}
\newcommand{\zkn}[1]{\todo[color=orange!20, size=\tiny]{#1}}
\newcommand{\uj}[1]{{\color{blue}{(\textbf{UJ: }#1)}}}
\newcommand{\ujn}[1]{\todo[color=blue!20, size=\tiny]{UJ: #1}}
\newcommand{\rdn}[1]{\todo[color=orange!20, size=\tiny]{RD: #1}}

\renewcommand{\zk}[1]{}
\renewcommand{\zkn}[1]{}
\renewcommand\asz[1]{}
\renewcommand{\aks}[1]{}
\renewcommand{\rd}[1]{}
\renewcommand{\uj}[1]{}
\renewcommand{\ujn}[1]{}
\renewcommand{\rdn}[1]{}

\newcommand{\pbt}{PBT\xspace}
\newcommand{\pbts}{PBT-State\xspace}
\newcommand{\xp}{PBT\xspace}
\newcommand{\xps}{PBT (GT State)\xspace}
\newcommand{\xporacle}{GT Coord\xspace}
\newcommand{\splay}{SP\xspace}
\newcommand{\splays}{SP (GT State)\xspace}
\newcommand{\fcp}{FCP\xspace}
\newcommand{\tdi}{TrajeDi\xspace}
\newcommand{\method}{Behavior Diversity Play\xspace}
\newcommand{\met}{BDP\xspace}
\newcommand{\metxpse}{BDP - [Discrim, Latent]\xspace}
\newcommand{\metpr}{BDP - [Discrim]\xspace}
\newcommand{\metxd}{BDP - [Latent (Sep Enc)]\xspace}
\newcommand{\metxdse}{BDP - [Latent (Shared Enc)]\xspace}

\newcommand{\bg}{behavior generator policy\xspace}
\newcommand{\Bg}{Behavior generator policy\xspace}
\newcommand{\BG}{Behavior Generator Policy\xspace}

\newcommand{\StageA}{Learning Behavior Policy\xspace}
\newcommand{\stagea}{learning behavior policy\xspace}
\newcommand{\StageB}{Learning the Coordination Policy\xspace}
\newcommand{\stageb}{learning the coordination policy policy\xspace}
\newcommand{\xhdr}[1]{\vspace{2pt}\textbf{#1}}

\newcommand{\settable}{Set Table\xspace}
\newcommand{\tdyhouse}{Tidy House\xspace}
\newcommand{\sourcedest}{https://bit.ly/43vNgFk}
\newcommand{\sourcedestshort}{https://bit.ly/43vNgFk}
\newcommand{\prepgroc}{Prepare Groceries\xspace}

\twocolumn[
\icmltitle{Adaptive Coordination in Social Embodied Rearrangement}

\icmlsetsymbol{equal}{*}

\begin{icmlauthorlist}
\icmlauthor{Andrew Szot}{Meta,GT}
\icmlauthor{Unnat Jain}{Meta}
\icmlauthor{Dhruv Batra}{Meta,GT}
\icmlauthor{Zsolt Kira}{GT}
\icmlauthor{Ruta Desai}{Meta}
\icmlauthor{Akshara Rai}{Meta}
\end{icmlauthorlist}

\icmlaffiliation{GT}{Georgia Institute of Technology}
\icmlaffiliation{Meta}{Meta AI}

\icmlcorrespondingauthor{Andrew Szot}{aszot3@gatech.edu}

\icmlkeywords{Machine Learning, ICML}

\vskip 0.3in
]

\printAffiliationsAndNotice{}  %

\begin{abstract}
\looseness=-1 
We present the task of ``Social Rearrangement", consisting of cooperative everyday tasks like setting up the dinner table, tidying a house or unpacking groceries in a simulated multi-agent environment. In Social Rearrangement, two robots coordinate to complete a long-horizon task, using onboard sensing and egocentric observations, and no privileged information about the environment. We study zero-shot coordination (ZSC) in this task, where an agent collaborates with a new partner, emulating a scenario where a robot collaborates with a new human partner. Prior ZSC approaches struggle to generalize in our complex and visually rich setting, and on further analysis, we find that they fail to generate diverse coordination behaviors at training time. 
To counter this, we propose Behavior Diversity Play (BDP), a novel ZSC approach that encourages diversity through a discriminability objective. 
Our results demonstrate that BDP learns adaptive agents that can tackle visual coordination, and zero-shot generalize to new partners in unseen environments, achieving $ 35\% $ higher success and $ 32\%$ higher efficiency compared to baselines.
\end{abstract}

\section{Introduction}

\begin{figure*}[t] 
  \centering
  \includegraphics[width=\textwidth]{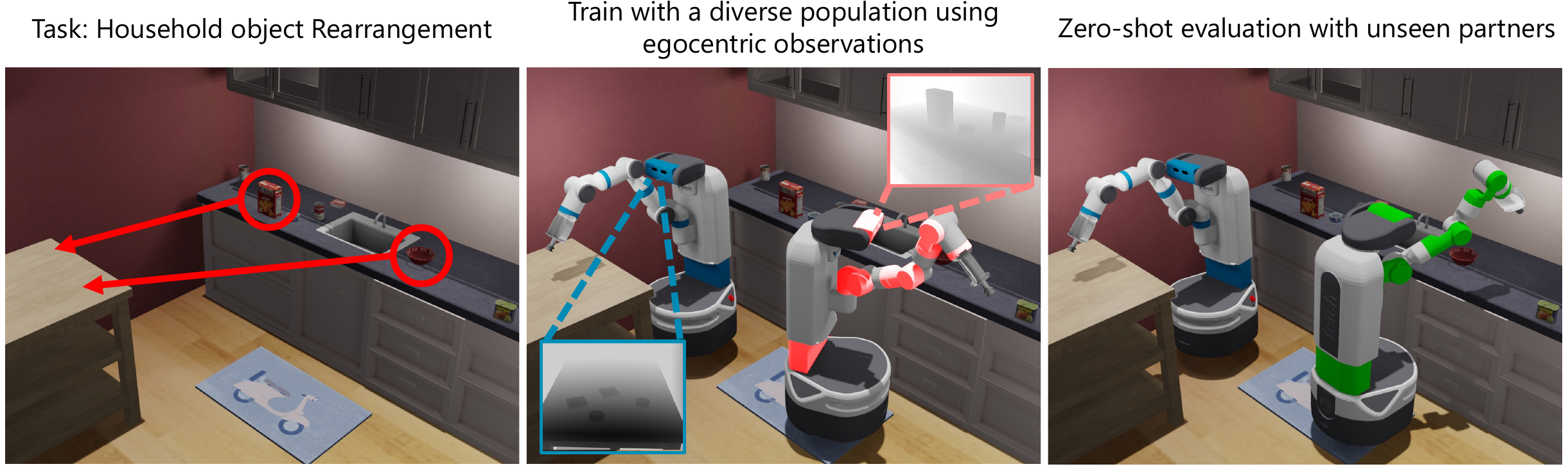}
  \caption{ 
    \textbf{Overview}. (Left) Task: Rearranging two objects from a start to a goal location. 
    (Middle) The blue robot learns to coordinate to rearrange the objects as efficiently as possible, with diverse red partners. The robots operate from egocentric visual and proprioceptive observations.
    (Right) The blue robot now coordinates with an unseen green robot \emph{zero-shot}.
    }
  \label{fig:teaser}
  \vspace{-0.5cm}
\end{figure*}

Consider a human-robot or robot-robot team, collaborating at everyday tasks like unloading groceries, preparing dinner or cleaning the house. 
Such an assistive robot should coordinate with its partner to efficiently complete the task, without getting in their way. 
For example, while tidying the house, if its partner starts cleaning the kitchen, the robot could start cleaning the living room to maximize efficiency. 
If the robot notices its partner loading the dishwasher, it should prioritize bringing dirty dishes from the living room to the kitchen, instead of rearranging cushions. 
The robot should be able to reason about its embodiment to avoid getting in the way of its partner while acting to effectively assist them. 
There are several challenges in building such a collaborative system.
(1) The robot needs to adapt to preferences of its partner, which might be unobserved, and change over time. 
For example, its partner might do a different part of the task in different situations, and the robot must adapt to such changes. 
(2) The environment and the partner are partially observed through the robot's egocentric cameras, making both inferring the state of the partner and the environment challenging. 
(3) The tasks are complex and long-horizon, with feasibility constraints that affect both the robot's and its partner's actions. 
For example, once the robot infers that its partner is loading dishes, it must bring dishes to the kitchen to enable its partner to succeed. 
All of these challenges make multi-agent collaboration in visually-realistic, long-horizon tasks challenging.

Zero-shot coordination (ZSC) \cite{lanctot2017unified, strouse2021collaborating} -- a two-stage learning framework that first trains a diverse population of agents (typically enforced through random policy initializations), and then trains a \cagent to collaborate with this population -- has been used to study such problems. 
However, so far ZSC approaches have only been applied to simplistic environments and tasks, with complete (privileged) information, like Overcooked~\cite{carroll2019utility}. Instead, real-world coordination requires dealing with partial information, and high-dimensional, continuous observations like images. Such a visually-rich setting requires bulky policy architectures, and the naive strategy of random policy initializations for generating different behaviors~\cite{strouse2021collaborating} is not enough. As a result, we observe that most agents in the population exhibit similar behaviors, like always reaching for the bowl when setting the dinner table. A \cagent trained with such a population is not adaptive to other partner preferences, like reaching for the fruit. To solve this problem, we propose a novel approach for ZSC -- Behavior Diversity Play (BDP) -- which uses a shared policy architecture and a discriminability objective to encourage behavioral diversity. Specifically, we train a discriminator network to distinguish the population behaviors given a history of states, encouraging the population behaviors to be distinct. For example, when setting a dinner table, different agents in the population attempt to do different parts of the task, like reach for the fruit or the bowl. A \cagent trained with this population is robust to either behavior at test-time, making it adaptive to unseen partners. Moreover, our shared policy architecture allows sample-efficient learning of bulky visual encoders, and parameterizes populations using a behavior latent space instead of random policy initialization. This architecture makes ZSC scalable for multi-agent collaboration in realistic, high-dimensional environments.

Our second contribution is a realistic multi-agent collaboration environment ``Social Rearrangement", consisting of everyday tasks like setting up the dinner table, tidying a house or unpacking groceries. 
Social Rearrangement is simulated in AI Habitat~\cite{habitat19iccv, szot2021habitat} -- a high-throughput physics-enabled photo-realistic 3D simulator. Two Fetch robots \cite{fetchrobot} are instantiated in a fully-furnished apartment from the ReplicaCAD dataset \cite{szot2021habitat}, and tasked with solving everyday, long-horizon tasks (see \Cref{fig:teaser}). The robots do not have access to any privileged information, like a bird's eye view of the house, or actions of their partner, and must operate entirely from onboard camera and proprioceptive sensing. We treat robot-robot cooperation as a proxy for human-robot cooperation and thus, don't assume access to the unobserved preferences or inner workings (policies) of the partner. The agent must coordinate with new partners, as observed through its egocentric cameras. Unseen partners have (scripted or learned) task-specific preferences, some might do only one part of the task, and others might do nothing; the \cagent must adapt to this range of behavior. 

We evaluate ZSC at Social Rearrangement, and observe that state-of-the-art ZSC approaches have poor generalization performance in this environment, due to a lack of behavior diversity in their learned population. 
Instead, \met learns diverse coordination behaviors in its population, with the help of the discriminability objective which encourages agents in the population to exhibit distinct behaviors, and in turn, can be used to train an adaptive \cagent. 
Our experiments show \met achieves $ 35\%$ higher success and is $ 32\% $ more efficient when coordinating with unseen agents compared to the approach from \cite{jaderberg2019human}, averaged over 3 tasks.
Finally, we present approaches for analysing population behavior and diversity, and show that higher diversity results in stronger zero-shot coordination. 

Overall, the key contributions of our work are: (1) We propose a novel ZSC approach -- Behavior Diversity Play (BDP) which outperforms prior ZSC approaches at visually-rich tasks. (2) We present the Social Rearrangement task for collaborative embodied AI research, featuring realistic everyday home tasks like tidying up a house. (3) We benchmark ZSC approaches at complex, long-horizon tasks, against unseen learned and scripted agents over 10,000 rearrangement problems in 60 environments. 
All code is available at \href{\sourcedest}{\sourcedestshort}.

\section{Related Work}

\textbf{Visual Embodied Agents.}
Embodied AI has seen great advancements in simulation platforms \cite{ai2thor,chang2017matterport3d,xia2018gibson,habitat19iccv,xia2019interactive,AllenAct,sapien,puig2018virtualhome,szot2021habitat} and new task specifications \cite{habitat19iccv,anderson2018vision,batra2020objectnav,chattopadhyay2021robustnav,chaplotNeuralTopologicalSLAM2020,wani2020multion,chen2019audio,gan2021threedworld}.
Object rearrangement, where a robot must interact with the environment to achieve a desired environment configuration is an important task for home robotics \cite{batra2020rearrangement}, and a variety of simulators support it \cite{weihs2021visual,shridhar2020alfred,teach,ehsani2021manipulathor,szot2021habitat}.
We utilize the Home Assistant Benchmark (HAB) in AI Habitat proposed by \cite{szot2021habitat}, consisting of home tasks like ``tidy the house", ``set the table", and ``prepare the groceries". Our \tname task extends the HAB to multi-agent setting. 

\textbf{Visual Deep Multi-Agent RL.}
Multi-agent RL (MARL) deals with learning policies for multiple embodied agents act to complete a task e.g., synchronized moving of furniture, that necessarily requires two agents~\cite{jain2019CVPRTBONE,JainWeihs2020CordialSync}.
Beyond collaborative tasks, competitive tasks like hide-and-seek and soccer have been investigated~\cite{chen2019visual,juliani2018unity,weihs2021learning,GoogleResearchFootball}. Visual MARL has been studied in the heterogeneous setting -- where embodied agents have different capabilities~\cite{thomason2020vision,rmm,patel2021comon} and teacher-student framework~\cite{WeihsJain2020Bridging,jain2021gridtopix}. Visual MARL has also been deployed to realistic, and procedurally-generated abstract environments~\cite{jaderberg2019human,team2021open}. 
Prior work on committed exploration for MARL~\cite{mahajan2019maven} shows basic adaptation to structural changes in environment and task setup.
Building on, but the above works, we learn agents that can adapt to \emph{novel partners} at evaluation time. 
We make the choice to not model communication between the agents and study coordination purely based on observing the partner.

\textbf{Adaptability in Multi-Agent RL.}
Ad-hoc teamwork~\cite{stone2010ad,barrett2011empirical} studies how agents can adapt their behavior to join teams. 
Similarly, theory of mind~\cite{premack1978does} studies how modeling the behavior of partner agents improves coordination robustness~\cite{choudhury2019utility,sclar2022symmetric}.
Previous works~\cite{puig2020watch,carroll2019utility} study how to coordinate with humans, but assume  privileged information about the partner in the form of a learned or planner-based explicit model of the partner.
The related problem of \emph{zero-shot coordination} (ZSC) \cite{hu2020other} studies how agents generalize to new partners, without any fine-tuning. Overcooked~\cite{carroll2019utility} is a simulated, simplified kitchen benchmark for studying ZSC with a discrete state and action space. 
Hanabi~\cite{bard2020hanabi} is another common benchmark for ZSC~\cite{hu2020other,hu2021off,lupu2021trajectory}.
In contrast to these low-dimensional environments, we study ZSC in \tname, a complex, visually-realistic 3D environment.
\cite{charakorn2020investigating,mckee2022quantifying} show that multi-agent RL benefits from diversity over partners and environments. We address diversity through a novel ZSC approach and benchmark it over 10,000 different rearrangement problems in 60 environments. %

\textbf{Zero-shot coordination (ZSC).} Some ZSC methods rely on known symmetries in the environment~\cite{hu2020other}, known environment models~\cite{hu2021off}, or simplified state spaces for manually defining coordination events~\cite{wu2021too}. 
Alternatively, population-play~\cite{jaderberg2017population} is a two-stage ZSC framework which first trains a population of agents through random pairing, and next trains an agent to coordinate with all agents in the population. Such approaches strive for diverse policy distributions at train time~\cite{heinrich2016deep,heinrich2015fictitious}, which results in an adaptable \cagent (sometimes called the ``best response policy'' in this literature). 
Fictitious co-play \cite{strouse2021collaborating} extends this by incorporating previous checkpoints in the population to represent varied skill levels.
These approaches rely on random network initializations and stochastic optimization to achieve behavior diversity, which is not sufficient for diversity in our tasks. Other works introduce auxiliary diversity objectives based on action distributions~\cite{lupu2021trajectory,zhao2021maximum,rahman2022towards}, which are also not well-suited to embodied tasks where different actions can lead to the same states.
Instead, we use a discriminability-based diversity objective, conditioned on a history of states, with a new policy architecture to aid learning in visual environments. 
Specifically, we share policy parameters between agents of a population, to enable scalable, sample-efficient population training, and parametrize the population using a behavior latent space. 

\textbf{Modeling Diverse Behaviors}.
Akin to \emph{quality diversity}~\cite{pugh2016quality, cully2015robots, krause2014submodular}, our method learns policies that are diverse and proficient at rearrangement.
Prior work has explored low-dimensional latent spaces for behaviors~\cite{derek2021adaptable}, though not in the context of ZSC.
DIAYN~\cite{eysenbach2018diversity} learns diverse skills using an unsupervised objective in single-agent settings.
MAVEN~\cite{mahajan2019maven} uses latent spaces to learn diverse exploration strategies in a multi-agent setting.
\cite{wang2022co} use a latent space to learn diverse behaviors from a multi-agent dataset.
We also use a behavior latent space, and policies conditioned on this space, but using RL, and in the context of ZSC.

\section{\tname}

We introduce the task of \emph{\tname} where two Fetch robots \cite{fetchrobot} solve a long-horizon everyday task (like tidying a house) in a realistic, visual 3D environment. 
While both agents work together, they do not know each other's policy, similar to how assistive robots must adapt to their partner's behavior.
The robots also do not have access to any privileged information (like a bird's eye view of the house or actions of the partner) and must operate entirely from an onboard egocentric camera and proprioceptive sensing.
At evaluation time, learned agents coordinate with new partners in new environments with new object placements and furniture layout. 
This emulates realistic collaboration, where two agents complete a rearrangement task together, while implicitly inferring their partner's state to aid or avoid getting in each other's way.
This is a significantly more complex setting than previous environments used to test multi-agent collaboration, like Overcooked \cite{carroll2019utility}, which assumes privileged information about the environment (top-down map) and operates in a low-dimensional, discrete state and action space.

In \tname, the agents must move $N$ objects from known start to end positions, both specified by 3D coordinates in each robot's start coordinate frame.
We build on the Home Assistant Benchmark~\cite{szot2021habitat} in the AI Habitat simulator~\cite{habitat19iccv} that studies Rearrangement~\cite{batra2020rearrangement}. \tname extends Rearrangement to a collaborative setting, where two agents coordinate to rearrange objects as efficiently as possible. 
Both Fetch robots are equipped with the same observation space: (1) a head mounted depth camera with $ 90^{\circ}$ FoV and $ 256 \times 256$ pixel resolution, (2) proprioceptive state measured with arm joint angles and (3) base egomotion (providing relative distance and heading since the start of the episode, sometimes called GPS+Compass). 
Agents can also sense the relative distance and heading to their partner. 
Note that this does not reveal the partner's actions and intents, or even the partner's full state, like their arm joint angles or visual observations, but enables the agents to learn to avoid collisions, etc. 
Additionally, each agent receives the distance and heading to the target objects' start and desired end positions, as a way of specifying the task. If objects are in a closed receptacle, like a drawer or fridge, the robot needs to reason that it must first open the receptacle, before picking the object. Both agents move their base through linear and angular base velocity (2D, continuous) and move their arm by setting desired delta joint angle (7D, continuous).
Grasping is controlled by engaging a suction gripper when in contact with an object (1D, binary).

The agents are rewarded for completing the task in as few simulation steps as possible, and if they collide, the episode ends with failure. 
\tname consists of the following three tasks adapted from \cite{szot2021habitat}:
\begin{itemize}[itemsep=0pt,topsep=0pt,parsep=0pt,partopsep=0pt,parsep=0pt,leftmargin=*]
  \item \noindent  \textbf{Set Table:} Move a bowl from the fridge to the table, and place a fruit from the fridge in the bowl. 
    Both the fridge and drawer are initially closed, and they must be opened before removing the objects inside. 

  \item \noindent  \textbf{Tidy House:} Move two objects from initial locations to target locations. 
    The objects are spawned across six open receptacles, and assigned a goal on one of the 6 receptacles, different from the starting receptacle. 

  \item \noindent  \textbf{Prepare Groceries:} Move one object from an open fridge to the counter and another from the kitchen to the fridge.
\end{itemize}
The different tasks elicit different coordination strategies.
For example, in \settable an agent might prefer to always pick the bowl or always pick the fruit. 
On the other hand, when tidying the house an agent might be indifferent to the object type and always tidy the closest object first. 
These everyday tasks study the ability of coordination agents to perform complex, long-horizon tasks with unseen partners and realistic sensing.
While all tasks can be completed by a single agent, coordinating with a team would result in improved efficiency by dividing up the task. 
We follow the standard dataset split in the ReplicaCAD~\cite{szot2021habitat} scene dataset with YCB objects~\cite{calli2015ycb}; agents are evaluated in new layouts of the house with new object placements. 
We train and evaluate policies for each task independently. 
Details about the tasks are in \cref{sec:further_task_details}.

\begin{figure*}[!h]
  \centering
  \includegraphics[width=\textwidth]{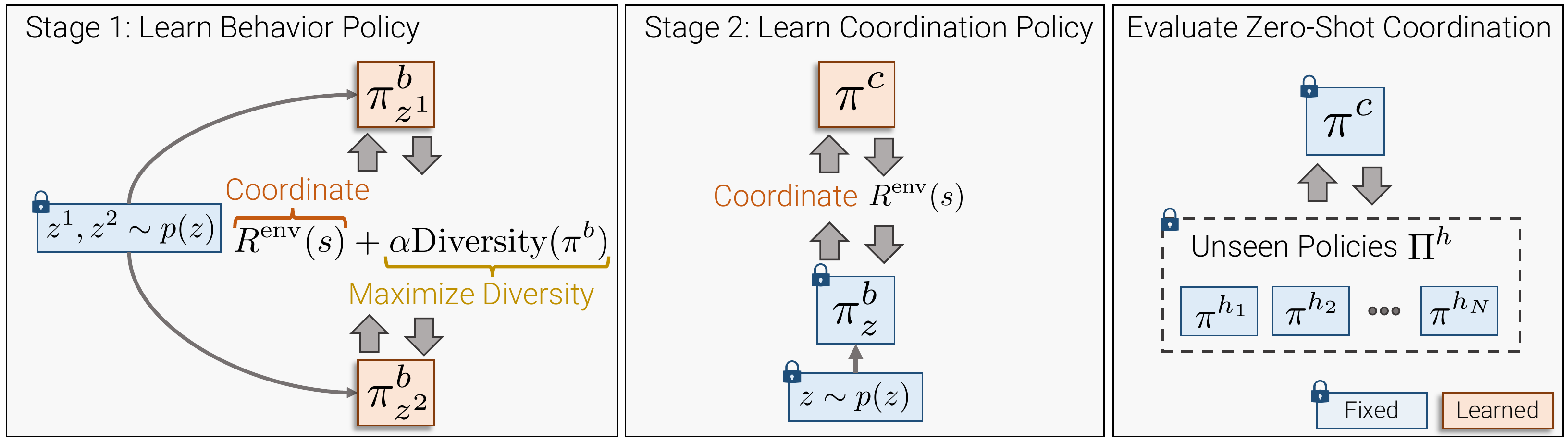}
  \caption{
    \textbf{\method}. (Left) Stage 1: We train the behavior policy $\pi^b$, which models a diverse set of agent behaviors, conditioned on the behavior latent $ z$. A discriminator $ q_\phi$ then encourages distinguishability between different $z$. 
    (Middle) Stage 2: A \cagent $ \pi^c$ learns to coordinate with different behaviors generated by the behavior policy.
    (Right) Stage 3: We evaluate the \cagent at ZSC with unseen holdout agents $\pi^h$.
  }
  \label{fig:method}
  \vspace{-0.5cm}
\end{figure*}

\section{Behavior Diversity Play}

Collaboration in everyday tasks requires adapting to unseen partner behaviors. 
For example, a partner may choose to do different parts of a task in different episodes and the learned `coordination agent' should adapt to such variations.
We propose a new method called \method (\met) which enables zero-shot coordination (ZSC) with unseen partners. 
Like prior ZSC methods,
\met consists of a two-stage training framework illustrated in \Cref{fig:method}. 
It first learns to generate task-relevant diverse behaviors and then trains the \cag to coordinate with these diverse partners. 
By coordinating with diverse partners at training time, the \cagent can generalize to unseen partners at test time. 
Specifically, in the first stage, we train a behavior policy generator $ \pi^{b}$ capable of generating diverse behaviors (\cref{fig:method}, left). 
Next, the \cag $\pi^c$ is trained to coordinate with the diverse behaviors generated by $\pi^b$ (\cref{fig:method}, middle), and finally evaluated against unseen holdout policies $\Pi^h$ (\cref{fig:method}, right). 
We first introduce our notation and a formal description of ZSC.
Next, we detail the two stage training process and finally provide practical details on implementing \met.

\met learns to generate diverse behaviors through a single behavior generator $ \pi^{b}$ by conditioning on different behavior latents, as opposed to prior works that learn independent policies.
This choice is conducive to our visually-rich setting where policy architectures can be large, slow to train, and difficult to fit multiple on GPUs. 
$\pi^b$ shares weights across different population agents, increasing training efficiency. Moreover, unlike prior work~\cite{strouse2021collaborating}, \met does not rely on random initialization of policies to ensure diverse behaviors, and instead incorporates an explicit diversity objective through a discriminator. This allows \met to train adaptive coordination agents across multiple tasks and experimental scenarios.

\subsection{Background and Notation}
\label{sec:notation} 
The goal of ZSC is to produce a \emph{\cag} $ \pi^{c}$ that can coordinate with unseen partners. 
For a given task, the \cag is evaluated based on zero-shot performance when paired with agents from an unseen, \textit{holdout policy set}, $ \Pi^{h}$.
Crucially, the policies in $\Pi^{h}$ are never seen during training. 
$ n$ agents solve the \tname task, which we formulate using a decentralized partially-observed markov decision process (Dec-POMDP) consisting of the tuple $\langle \mathcal{S}, \mathcal{A}, \mathcal{O}, \mathcal{R}, \mathcal{P}, \gamma, n, T \rangle$. 
In this work we use $ n =2$ agents, but our approach remains unchanged for $n>2$.
At each time step $t$, the global environment state is denoted as $s_t\in\mathcal{S}$. 
Each agent $ i$, receives an observation $o^i_t\in\mathcal{O}$, takes an action $a^i_t \in \mathcal{A}$, forming a joint action $\bm{a}_t \in \mathcal{A}^{n}$, resulting in the next environment state $s_{t+1}$, following the transition function $\mathcal{P}$. 
This gets joint reward $r_t \in \mathbb{R}$ according to the deterministic reward function $\mathcal{R}: \mathcal{S} \times {\cal A}^n \rightarrow \mathbb{R}$ for an episode of length $T$. 
Agent $ i$'s policy $ \pi^{i}$ maps its observations to a distribution over actions. 
To handle partial observability and history, policies are modeled with recurrent neural networks.
The combined expected return of policies $ \pi^1, \pi^2$, is $ J(\pi^1, \pi^2) = \sum_{t=0}^{T} \mathbb{E}_{\bm{a_t} \sim (\pi^1,\pi^2)} \left[ \gamma^{t}  \mathcal{R} (s_t, \bm{a_t}) \right] $.
The goal of the coordination agent $\pi^{\text{c}}$ is to maximize the average performance over the holdout population $\Pi^h$, $ \mathbb{E}_{\pi^{h} \sim \Pi^{h}}\left[J(\pi^{\text{c}}, \pi^{h})\right]$.
\vspace{-5pt}
\subsection{Stage 1: Behavior Policy Generator}
\vspace{-5pt}
We model diverse behaviors through a \emph{Behavior Policy Generator} policy $\pi^{b}(a_t | o_t, z)$, which conditioned on a \emph{behavior latent} $z \sim p(z)$ generates distinct behaviors per $z$. The behavior prior $p(z)$ is modelled as a uniform categorical distribution with $K$ categories, where $K$ is a hyperparameter, analogous to a population size. We train $\pi^b$ to maximize a joint task and diversity objective that encourages agent trajectories to be distinct, while completing the task at hand. %
Specifically, our training objective consists of two components: (1) task performance, aimed to learn agents that can solve the rearrangement task, and (2) diversity -- a discriminator-based reward to encourage distinct behaviors per behavior latent. 
At the start of each episode, we sample two latents $z^1, z^2 \sim p(z)$, one for each partner. 
For brevity, we denote $ \pi^b (\cdot | \cdot , z)$ as $\pi^{b}_z $. 
Next, we optimize:
\begin{align}
  \max_{\pi^b} J(\pi^b_{z^1} , \pi^b_{z^2}) + \alpha \diversity \left( \pi^b \right)
\end{align} 
where $J$ is the return described in Sec.~\ref{sec:notation} and Appendix \ref{sec:further_task_details}. Diversity is a measure of how diverse of behaviors $\pi^b$ can produce, while $ \alpha$ is weights the diversity objective.

By increasing diversity, we reduce the conditional entropy of the latent $z$ given the state history, while also increasing the entropy of the policy. 
Minimizing the entropy of $z$ given the state history encourages $ z$ to be predictable from the agent's behaviors, making the different behaviors generated by $\pi^b_z$ distinct. 
Maximizing the entropy of the policy ensures that different $z$ are diverse enough to cover the space of possible behaviors. 
To optimize this objective, \met learns a \emph{trajectory discriminator} $q_\phi$ that predicts which behavior latent corresponds to an agent's trajectory. 
Since the discriminator is only used during training, it enjoys access to each agent's privileged state trajectories $\tau^i = {s_0^i, \dots s_T^i}$, which are not available during evaluation. 
Ideally, $q_\phi(z|\tau^i)$, should allocate high probability to behavior latent for agent $i$, \textit{i.e.}, $z^i\in\{1\dots K\}$. 
We adapt the skill diversity formulation from \cite{eysenbach2018diversity} for coordination by conditioning on state trajectories, instead of states: 
\vspace{-4pt}
\begin{align}
  &\diversity(\pi^b) = -\mathcal{H}\left( z | \tau \right) + \mathcal{H}(a | o, z) \nonumber \\ 
  &\ge \mathop{\mathbb{E}}_{z \sim p(z), \tau \sim \pi_z^{b}} \left[ \log q_\phi(z | \tau) \right] +  \mathcal{H}(a | o, z)
\vspace{-5pt}
\label{eq:diversity}
\end{align}
where $ \mathcal{H}( \cdot )$ is the entropy function and the second line gives the variational lower bound on the diversity objective. 
$\log q_\phi(z | \tau) $ is a discriminator loss that enforces distinguishability, and $\mathcal{H}(a | o, z)$ is an entropy objective that encourages the policies to cover a large space of behaviors.

Trajedi~\cite{lupu2021trajectory} proposes an approach that encourages distinct action distributions induced by different policies to achieve population diversity. Instead, we propose to measure diversity in terms of induced state trajectories, which are more indicative of behaviors than actions. For example, different actions can lead to the same state changes, and hence result in the same high-level behaviors.
See \Cref{sec:further_cmp} for a detailed description of the difference in the diversity objectives of \met and Trajedi.
\vspace{-5pt}
\subsection{Stage 2: \StageB}
\vspace{-5pt}
After stage 1, the behavior policy generator $\pi^b$ can generate diverse behaviors when conditioned on different latents $z$. 
Next, we train a new policy \cagent $ \pi^c( a | o)$ to coordinate with \textit{all} behaviors produced by $ \pi^b$ from the first stage (middle panel of \cref{fig:method}), while keeping $\pi^b$ fixed. $ \pi^c$ is initialized randomly,
and then trained to maximize task performance $J(\pi^c, \pi^b_z)$ when paired with $ \pi^b_z$ for a randomly-sampled $ z$. 
It is hard for $ \pi^{c}$ to learn against a rapidly changing $ z$ as the behaviors it is paired with change constantly.
For this reason, we sample a new $ z$ only once every several updates of $ \pi^{c}$.
Exact training details for both stages are in \Cref{sec:method_details}.

Once trained, $ \pi^c$ is able to adapt to new partners zero-shot since it was trained to coordinate with the diverse behaviors from $ \pi^b$ (right panel of \cref{fig:method}). 
Next, it is evaluated against unseen partners from the holdout set $\Pi^h$ to measure its generalization performance. 

While we describe the above sections assuming $n=2$ agents collaborating, the approach remains unchanged for $n>2$ agents. For $n>2$, during Stage 1 training, BDP would sample $n$ agents instead of 2 and use the same diversity objective (Eq \ref{eq:diversity}). In Stage 2 training, the \cagent would learn to collaborate with $n-1$ partners generated by the behavior policy.

\subsection{Implementation Details}
\label{sec:impl_details} 

We use a two-layer hierarchical policy architecture for all baselines, where a high-level policy selects a low-level skill to execute based on observations. This has shown to be effective in rearrangement tasks~\cite{gu2022multi}. 
We consider a known, fixed library of low-level skills, which can achieve instructions like `navigate to the fridge', or `pick an apple'. 
These low-level skills directly interact with the environment via low-level base and arm actions.
The action space of the learned high-level policy is a discrete selection over all possible combinations of skill and object/receptacle parameterizations allowed at all steps.
For navigation, we allow all possible furniture pieces to be navigable to, and manipulation skills can interact with any articulated receptacles, target objects, and goals.
We also include additional high-level navigation actions, like move-forward, turn-left, turn-right, and no-op in the action space that facilitate coordination between agents, like move out of the way, if it sees its partner coming towards itself to avoid collision. 

If the policy chooses to execute an infeasible action, like pick an object when its not within reach (based on hand-defined pre-conditions), the action results in a no-op and no change to the environment. 
Since the focus of our work is on high-level coordination, we assume access to perfect low-level skills for all approaches.
For manipulation skills, this means kinematically applying the hand-defined post-conditions of the skill, like attaching target objects in the scene to the gripper after executing the pick skill. %
For the navigation skill, we use a shortest path navigation module which moves the robot from its current position to the desired position (such as a receptacle or object), but does not take the partner agent into account. 
Avoiding collisions and coordination with the partner are dealt with by the high-level policy. 
More details on the hierarchical policy, policy architecture, discriminator, and pseudocode are in \Cref{sec:method_details}. While we use a hierarchical policy architecture, BDP itself is agnostic to the policy type. Prior works like \cite{szot2021habitat} have shown that such a policy architecture is well-suited to learning long-horizon rearrangement tasks, and hence the architecture used in our work.
Additionally, we make some simplifications to our simulation environment by using a partially simulated physics engine that considers collisions with objects like tables and the partner agent, but ignores collisions with the fridge door. On the other hand, aspects that are important for coordination, like colliding with the partner, are fully simulated. More details on the simulation can be found in Supp. \ref{sec:further_task_details}.

\begin{table*}[ht] 
  \centering
  \resizebox{\textwidth}{!}{
    \begin{tabular}{raaccccb}
      \toprule
       & \textbf{\pbts} & \textbf{\xporacle} & \textbf{\splay} & \textbf{\xp} &  \textbf{\fcp} & \textbf{\tdi} & \textbf{\met } \\
       & \textit{(\Oracle, No Vision)} & \textit{(\Oracle, No ZSC)} & \cite{heinrich2016deep} & \cite{jaderberg2019human} & \cite{strouse2021collaborating}  & \cite{lupu2021trajectory} &  \textit{(Ours)} \\
      \midrule
      \multicolumn{7}{@{}l}{\textbf{Set Table}}\\
      \traintitle &  70.74 {\scriptsize$ \pm$ 0.05}   &  90.52 {\scriptsize$ \pm$ 0.05}   &  57.74 {\scriptsize$ \pm$ 0.01}   &  46.67 {\scriptsize$ \pm$ 0.02}   &  29.90 {\scriptsize$ \pm$ 0.07}   & 43.24 {\scriptsize $ \pm$ 0.09} &  74.81 {\scriptsize$ \pm$ 0.01}   \\
\evaltitle &  50.39 {\scriptsize$ \pm$ 0.09}   & - &  17.97 {\scriptsize$ \pm$ 0.04}   &  30.34 {\scriptsize$ \pm$ 0.04}   &  37.50 {\scriptsize$ \pm$ 0.04}   & 32.52 {\scriptsize $ \pm$ 0.04} &  \textbf{46.43 {\scriptsize$ \pm$ 0.08}} 
 \\
      \midrule
      \multicolumn{7}{@{}l}{\textbf{Tidy House}}\\
      \traintitle &  74.90 {\scriptsize $ \pm$ 21.59  } &  92.28 {\scriptsize $ \pm$ 1.66  } &  34.18 {\scriptsize $ \pm$ 6.05  } &  36.13 {\scriptsize $ \pm$ 0.98  } &  12.04 {\scriptsize $ \pm$ 2.28  } &  39.65 {\scriptsize $ \pm$ 0.59  } &  73.83 {\scriptsize $ \pm$ 7.03  } \\
\evaltitle &  68.08 {\scriptsize $ \pm$ 0.09  } & - &  52.67 {\scriptsize $ \pm$ 0.06  } &  56.88 {\scriptsize $ \pm$ 0.07  } &  34.07 {\scriptsize $ \pm$ 0.09  } &  63.58 {\scriptsize $ \pm$ 0.05  } & \textbf{  66.71 {\scriptsize $ \pm$ 0.05  } }
 \\
      \midrule
      \multicolumn{7}{@{}l}{\textbf{Prepare Groceries}}\\
      \traintitle &  85.74 {\scriptsize $ \pm$ 2.82  } &  93.63 {\scriptsize $ \pm$ 0.28  } &  47.07 {\scriptsize $ \pm$ 29.88  } &  69.34 {\scriptsize $ \pm$ 1.76  } &  44.40 {\scriptsize $ \pm$ 6.38  } &  34.56 {\scriptsize $ \pm$ 27.94  } &  89.67 {\scriptsize $ \pm$ 2.51  } \\
\evaltitle &  77.01 {\scriptsize $ \pm$ 0.05  } & - &  56.04 {\scriptsize $ \pm$ 0.07  } &  56.08 {\scriptsize $ \pm$ 0.09  } &  30.00 {\scriptsize $ \pm$ 0.07  } &  53.84 {\scriptsize $ \pm$ 0.08  } & \textbf{  75.85 {\scriptsize $ \pm$ 0.05  } }
 \\
      \bottomrule
    \end{tabular}
  }
  \caption{
    Evaluation of \tname with training population and ZSC with unseen agents. \met outperforms prior ZSC works and closes the gap to \oracle\ methods (columns in gray). 
    Average and standard error across 3 seeds.
  }
  \label{tab:zsc} 
  \vspace{-0.5cm}
\end{table*}

\section{Experiments}
In this section, we compare our approach (\met) to state-of-the-art (ZSC) methods.
First, we introduce baselines, metrics, and quantitative results, then analyze the learned populations from the different approaches with a focus on measuring diversity. 
Lastly, we run ablations on \met to quantify the contribution of the policy architecture and the discriminator loss in the overall performance of \met.

\subsection{Baselines}
\label{sec:baselines}
We compare \method (\met) to a range of zero-shot coordination (ZSC) methods. 
\begin{itemize}[itemsep=0pt,topsep=0pt,parsep=0pt,partopsep=0pt,parsep=0pt,leftmargin=*]
  \item \textbf{Self-Play (\splay)}~\cite{heinrich2016deep} learns a population of size $N$ by randomly initializing $N$ policies and training them via self-play.

  \item \textbf{Population-Based Training (\xp)}~\cite{jaderberg2019human} initializes $ N$ random policies, and pairs them randomly during training.  
    Both \xp and \splay generate diversity through random policy initializations.

  \item \textbf{Fictitious Co-Play (\fcp)} \cite{strouse2021collaborating} uses SP but adds earlier checkpoints from each policy to the population when training the \cag.

  \item \textbf{Trajectory Diversity (\tdi) } \cite{lupu2021trajectory} adds a diversity objective to population training that encourages diverse action distributions as opposed to the diverse state distributions encouraged by \met.
\end{itemize}

Implementation details are in \Cref{sec:further_exp_details}. All of the above and \met follow two-stage training and only differ in how they obtain a population (the second stage of training the \cagent is identical between them).
The policy and environment setup for baselines is identical to the setup for \met described in \Cref{sec:impl_details}.
For \met, we model the behavior latent prior $ p(z)$ as a fixed 8-dimension uniform categorical distribution.
Likewise, all baselines train a population of $ 8$ agents .
Additionally, we implement two `\oracle' baselines that have privileged information to disentangle the two challenges of \tname  -- high-dimensional visual observations and  zero-shot coordination.
\begin{itemize}[itemsep=0pt,topsep=0pt,parsep=0pt,partopsep=0pt,parsep=0pt,leftmargin=*]
  \item \textbf{\pbts} (\Oracle, No-Vision) To highlight the challenges of operating from visual input, we implement \pbt with ground-truth environment state (\pbts), that captures the complete environment state (details of ground-truth state in \Cref{sec:gt_state}). 
\pbts operates in a fully observable and low-dimensional environment, similar to prior work like Overcooked~\cite{carroll2019utility}. 

  \item \textbf{\xporacle} (\Oracle, No-ZSC) To highlight the challenges of zero-shot coordination, we train two visual policies together (\xporacle). \xporacle operates on high-dimensional visual observations, but is trained together with its partner agent, and hence has no ZSC challenges.
\end{itemize}

\subsection{ZSC Evaluation}
\label{sec:eval-details} 
As introduced in~\cref{sec:notation}, to evaluate the \cagent $\pi^c$ trained using the different approaches, we task them to coordinate with a set of holdout agents $\Pi^h$ unseen during training. 
For each task, the holdout set consists of three scripted and eight learned holdout agents, described below. Further details of the holdout agents in \Cref{sec:further_zsc_details}.

\noindent\textbf{Scripted holdout agents} execute a fixed sequence of hard-coded task plans, for \textit{e.g.}, ``navigate to the table, pick up the object, navigate to the counter and drop the object", exhibiting different behavioral preferences. 
Note that the scripted holdout agents are not reactive, \textit{i.e.}, they will not update their plan based on the coordination agent's actions, or even try to avoid bumping into the coordination agent. 
This is out-of-distribution for the \cagent, since it is trained with partners that react to its actions, making coordinating with scripted holdout agents especially challenging. 

\noindent\textbf{Learned holdout agents} are separately trained using \xporacle and unseen during training. The \cagent needs to adapt to these unseen policies, which were trained to expect a particular behavior from their partner.

\noindent\textbf{Metrics.} For the three tasks -- Set Table,
Tidy House,
Prepare Groceries -- we compare the methods using the portion of successful task completions (1) when paired with agents from the training population (\textit{train population eval}, or \emph{\train} for short), (2) and ZSC with unseen agents from the holdout set (\textit{\eval}). We report mean and standard deviation across 3 randomly seeded runs calculated on 100 episodes in unseen scene configurations. 

\subsection{ZSC Quantitative Analysis}

\noindent\textbf{Evaluation metrics.} \met outperforms prior ZSC baselines (\splay, \pbt, \fcp, \tdi) across all \tname tasks when comparing both \train and \eval success rate (\cref{tab:zsc}).
For Set Table, \met improves \train success rate by 17\% (from \splay's 57.74\% $\rightarrow$ \met's 74.81\%). Looking closely, we find that \splay is unable to coordinate with holdout policies,achieving a low \eval success rate of 17.97\% (while \met can reach 46.63\%). Similar boosts in \eval success rate of 3.1\% (63.58\% $\rightarrow$ 66.71\%) and 19.8\% (56.08\% $\rightarrow$ 75.85\%) are observed in Tidy House and Prepare Groceries tasks, respectively

\noindent\textbf{\met bridges gap to \oracle methods.} Despite the use of privileged information and assumptions by \oracle~methods, \met achieves comparable ZSC performance as them. 
In \Cref{tab:zsc}, we see that \met can reason about partner and environment state from it's visual observations, and performs close to \pbts, \met's  46.43\% vs. \pbts's 50.39\% for \eval success rate in Set Table task. 
Unsurprisingly, there is indeed a scope for improvement in the overall coordination, since BDP's performance is still lower than \oracle \xporacle (\met's 74.81\% vs. \xporacle's 90.52\%).
\Cref{sec:extended_zsc} contains further analysis of ZSC performance by breaking down ZSC by holdout agent type.

\noindent\textbf{\met coordinates efficiently.} 
An interesting observation for some baselines, like \fcp at Set Table, is that their \eval success is higher than \train ($37.50\%$ vs. $29.90\%$). This is because some of the \eval agents are more adept at solving the tasks than the ones learned in the training population. As a result, it is important to look not just at the task success, but also the \emph{\cgain} of solving the task as a pair (Tab. \ref{tab:zsc_eff}). 
\begin{table}[h]
  \centering
  \resizebox{1.0\columnwidth}{!}{
    \begin{tabular}{raccccb}
      \toprule
       & \textbf{\pbts} & \textbf{\splay} & \textbf{\xp} &  \textbf{\fcp} & \textbf{\tdi} & \textbf{\met } (\textit{Ours}) \\
      \midrule
      \multicolumn{7}{@{}l}{\textbf{Set Table}}\\
      \traintitle & \textcolor{Green}{$+84\%$} & \textcolor{Green}{$+49\%$} & \textcolor{Green}{$+34\%$} & \textcolor{BrickRed}{$-28\%$} & \textcolor{Green}{$ +36\%$ } & \textcolor{Green}{$+51\%$} \\
\evaltitle & \textcolor{Green}{$+58\%$} & \textcolor{BrickRed}{$-30\%$} & \textcolor{BrickRed}{$-13\%$} & \textcolor{BrickRed}{$-7\%$} & \textcolor{BrickRed}{$-1\%$} & \textcolor{Green}{$+19\%$} \\
 \\
      \midrule
      \multicolumn{7}{@{}l}{\textbf{Tidy House}}\\
      \traintitle & \textcolor{Green}{$+85\%$} & \textcolor{Green}{$+6\%$} & \textcolor{Green}{$+10\%$} & \textcolor{BrickRed}{$-28\%$} & \textcolor{Green}{$+21\%$} & \textcolor{Green}{$+57\%$} \\
\evaltitle & \textcolor{Green}{$+57\%$} & \textcolor{Green}{$+16\%$} & \textcolor{Green}{$+24\%$} & \textcolor{BrickRed}{$-11\%$} & \textcolor{Green}{$+30\%$} & \textcolor{Green}{$+32\%$} \\
 \\
      \midrule
      \multicolumn{7}{@{}l}{\textbf{Prepare Groceries}}\\
      \traintitle & \textcolor{Green}{$+77\%$} & \textcolor{Green}{$+26\%$} & \textcolor{Green}{$+45\%$} & \textcolor{Green}{$+18\%$} & \textcolor{Green}{$+12\%$} & \textcolor{Green}{$+71\%$} \\
\evaltitle & \textcolor{Green}{$+52\%$} & \textcolor{Green}{$+22\%$} & \textcolor{Green}{$+21\%$} & \textcolor{BrickRed}{$-6\%$} & \textcolor{Green}{$+17\%$} & \textcolor{Green}{$+44\%$} \\
 \\
      \bottomrule
    \end{tabular}
  }
  \caption{
    \textbf{\CGain}: \met improves efficiency for both the training population and in ZSC compared to all baselines in all tasks. The oracle \pbts method takes privileged state information as input.
  }
  \label{tab:zsc_eff} 
\end{table}

To calculate the \cgain of ZSC methods, we compute the average number of steps it takes a single agent to solve a task and compare it to the average number of steps taken by 2 agent teams, where both agents can act every single step.
In \Cref{tab:zsc_eff}, we observe that ZSC baselines like \pbt can actually \textit{make the partner slower}, even if the task completion rate is high. On the other hand, \met improves the efficiency of unseen partners (-13\% using \pbt versus +19\% using \met on the Set Table task). 
Again, as compared to the oracle \pbts, we observe that \met has lower ZSC and train-pop efficiency, owing to not having complete state information, and pointing towards scope for improvement. This points towards scope for improvement, but also an interesting future direction where cooperation efficiency might be studied (and improved) over repeated interactions with the same partner.

\subsection{Qualitative Diversity Analysis}
Characterizing the behaviors of a population is challenging, and agent behavior itself consists of long trajectories and diverse interactions, making existing diversity metrics not as meaningful~\cite{mckee2022quantifying}. 
We present a qualitative diversity analysis approach, by pairing the \textit{same} agent with a population of agents. 
Specifically, we study the behavior of the learned \cagent when paired with agents from the training population, and holdout set. 
Ideally, the \cagent exhibits diverse behaviors during training, and adapts its behavior to unseen test partners. 

To do this, we define sub-goals that occur in the successful completion of a task, for example, to successfully complete \tdyhouse, agents must rearrange 2 objects.
We record the portion of these sub-goals completed by the \cag.
If the \cagent is biased towards only doing some parts of the task, it will fail to coordinate with partners who prefer the same portions of the task. 
Note that this hand-designed task decomposition is only used during evaluation, and not imposed during training, and the observed behaviors emerge solely through our diversity objective.
See \Cref{sec:qual_result_details} for more details. %

\begin{figure}
  \includegraphics[width=\columnwidth]{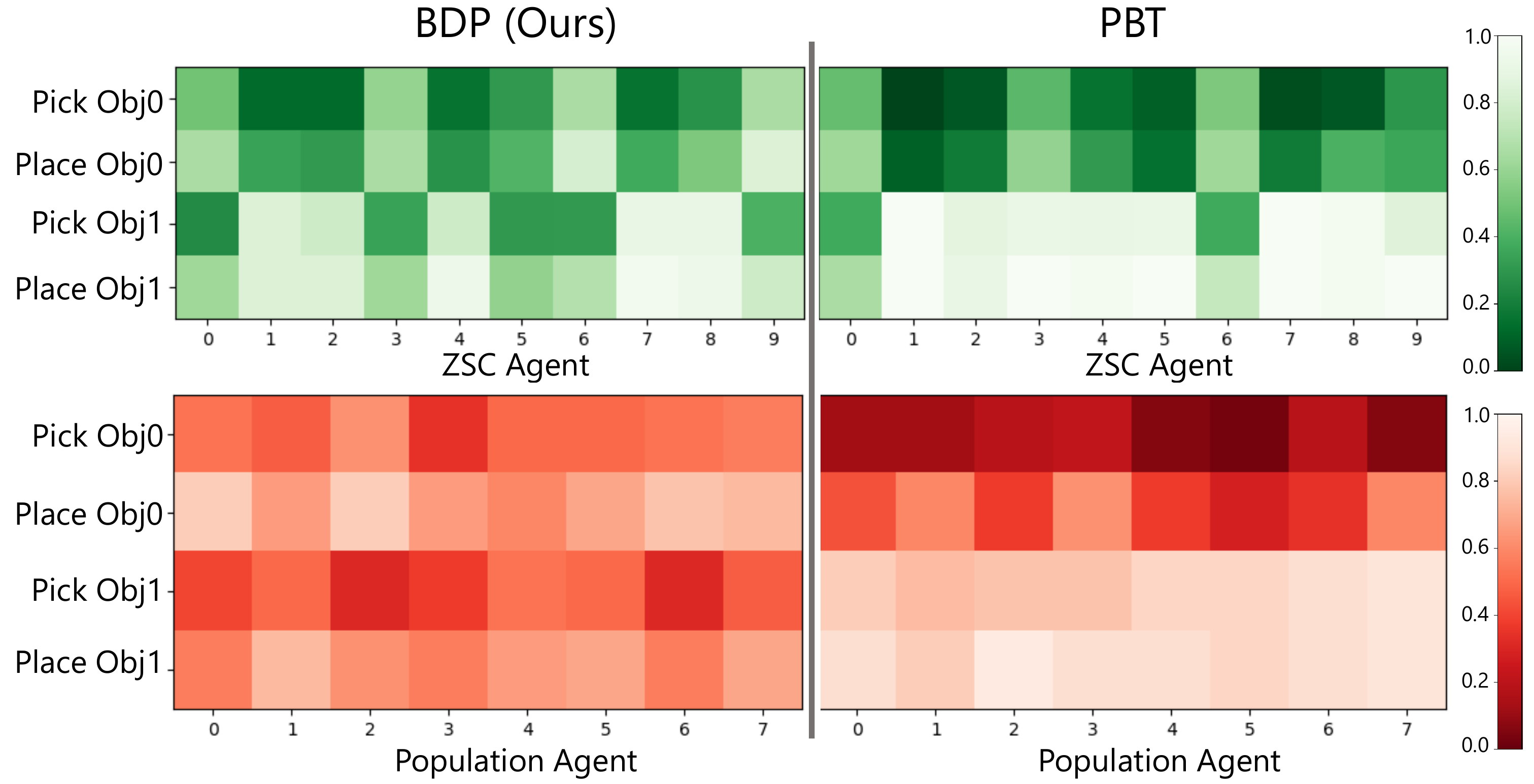}
  \caption{
    Behavior of the \cagent in the \tdyhouse task with \emph{unseen partners} during ZSC (top) and training population partners (bottom).
    Columns correspond to different partners for the \cag, rows are different sub-goals, and the cells display the probability of the sub-goal being completed by the \cagent.
  }
  \label{fig:zsc_study}
\end{figure}

\noindent\textbf{\met adapts to unseen partners.} \Cref{fig:zsc_study} (top) shows the probability of \cagent completing different sub-goals (rows), when paired with unseen holdout partners (columns). The \cagent trained with \met (\cref{fig:zsc_study}, top left) performs different portions of the task, depending on its partner. 
In contrast, \cagent trained with \pbt (\cref{fig:zsc_study}, top right) is biased towards a set of sub-goals, and hence can't coordinate with partners with the same bias. 

\noindent\textbf{\met results in diverse population.} Next, we study the training population trained in stage 1 by both \met and \pbt. 
Ideally, the \cag should exhibit diverse behaviors when paired with the training population, indicating that the population agents themselves have diverse behaviors and the \cag learns to adapt to them. 
In the bottom left of \Cref{fig:zsc_study}, we see that when trained with \met the \cagent is unbiased, and almost equally likely to perform any sub-goal, making it highly adaptable. 
In contrast, using \pbt, the \cagent only typically picks the first object (\cref{fig:zsc_study}, bottom right), implying that the training population from \pbt is biased towards picking the second object, and hence the \cagent does not see diverse behavior during training. %

\subsection{Ablation Experiments}
We ablate the new diversity objective in \met and show that removing the discriminator objective adversely impacts performance. 
Specifically, we implement the following variants of \met: 
\textbf{\metpr} keeps the shared network parameters and latent space, but we remove the discriminator objective. 
Diversity only comes from different input $ z$. 
\textbf{\metxpse} uses a shared visual encoder, but removes the latent space and diversity objective, instead using random initializations for diversity. 
Finally, \textbf{\pbt} has neither a shared latent, discriminator objective, or shared encoder.
The results in \Cref{tab:abl_discrim} show \metpr suffers in \eval (even though \train performance is high) since without the discriminator, there is nothing to enforce different latents to have different behaviors, making \eval poor. 
Both \metxpse and \xp rely on different network parameters to achieve behavior diversity, insufficient for our task, reinforcing the importance of a shared latent and discriminator objective.
\cref{sec:policy_arch_abl} contains policy architecture ablations.

\begin{table}[h] 
  \centering
  \resizebox{\columnwidth}{!}{
    \begin{tabular}{ccccb}
      \toprule
      & \textbf{\metpr} & \textbf{\metxpse} & \textbf{\xp}  & \textbf{\met} \\
\midrule
\traintitle &  77.18 {\scriptsize}$ \pm$ 0.00   &  42.16 {\scriptsize}$ \pm$ 0.01   &  46.67 {\scriptsize}$ \pm$ 0.02   &  74.81 {\scriptsize}$ \pm$ 0.01   \\
\evaltitle   &  22.92 {\scriptsize}$ \pm$ 0.05   &  33.20 {\scriptsize}$ \pm$ 0.04   &  30.34 {\scriptsize}$ \pm$ 0.04   &  46.43 {\scriptsize}$ \pm$ 0.08   
 \\
      \bottomrule
    \end{tabular}
  }
  \caption{
    Ablations on the new diversity objective in \met. 
  }
  \label{tab:abl_discrim} 
\end{table}

\vspace{-10pt}
\section{Conclusion}
We present the \tname task, consisting of collaborative, everyday tasks like tidying a house, setting a dinner table and preparing groceries. 
\tname is simulated using realistic, high-dimensional observations, with no privileged information like top-down maps, or partner actions. We present a novel approach \method (\met) for zero-shot coordination (ZSC) and evaluate it on \tname.
\met trains an adaptable \cagent that can collaborate with a set of unseen holdout policies, and improves the efficiency of its partner over solving the task alone. 
Through analysis and ablations, we show that this improvement comes from a diverse training population obtained via BDP's discrimibility objective.

While \met is able to learn adaptive agents that can use partial information about the environment and their partners to coordinate, its performance is worse than oracles with privileged state and partner information. This implies that there is still scope for improvement for \met at ZSC tasks. Furthermore, Social Rearrangement deals with a limited set of rearrangement tasks, with some simplifying assumptions like clean visual inputs and simplified physics. 

In the future, we hope to include more complex coordination tasks like furniture assembly or cooking a meal, which might even require additional inputs like language.
Future work can also improve BDP by exploring how theory of mind (ToM) can improve coordination by having the coordination policy predict the behavior policy's internal state or future actions.
Such a ToM objective can help BDP learn representations that predict the other agent's intentions, improving generalization. 
We hope that \tname can serve as a realistic test-bed for multi-agent collaboration research, and the approaches and analysis presented in our work can enable future research on ZSC. 

\section{Acknowledgments}
The Georgia Tech effort was supported in part by NSF, ONR YIP, and ARO PECASE. The views and conclusions contained herein are those of the authors and should not be interpreted as necessarily representing the official policies or endorsements, either expressed or implied, of the U.S. Government, or any sponsor.

{\small
\bibliography{egbib}
\bibliographystyle{icml2023}
}

\appendix

\section*{Appendix}
To view qualitative behavior, please view our supplemental video. 
We structure the Appendix as follows:
\begin{enumerate}[itemsep=2pt,topsep=0pt,parsep=0pt,partopsep=0pt,parsep=0pt,leftmargin=*]
    \item [\ref{sec:further_task_details}] Additional details about the \tname setup including reward structure.
    \item [\ref{sec:further_cmp}] Comparison of the diversity objective of the proposed \method and prior work of Trajedi~\cite{lupu2021trajectory}.
    \item [\ref{sec:method_details}] Necessary implementation details of \method for reproducibility -- training pipeline, policy architecture, and hyperparams.
    \item [\ref{sec:further_exp_details}] Further description of evaluations we conducted for comparing \method, four previous works, and privileged baselines.
    \item [\ref{sec:further_exps}] New experimental results including a breakdown of \Cref{tab:zsc}, more ZSC results, full results for \Cref{tab:zsc_eff}, and new ablations supplementing \Cref{tab:abl_discrim}. 
\end{enumerate}
The source code can be found at: \href{\sourcedest}{\sourcedest}

\begin{figure*}[!h]
  \centering
  \includegraphics[width=\textwidth]{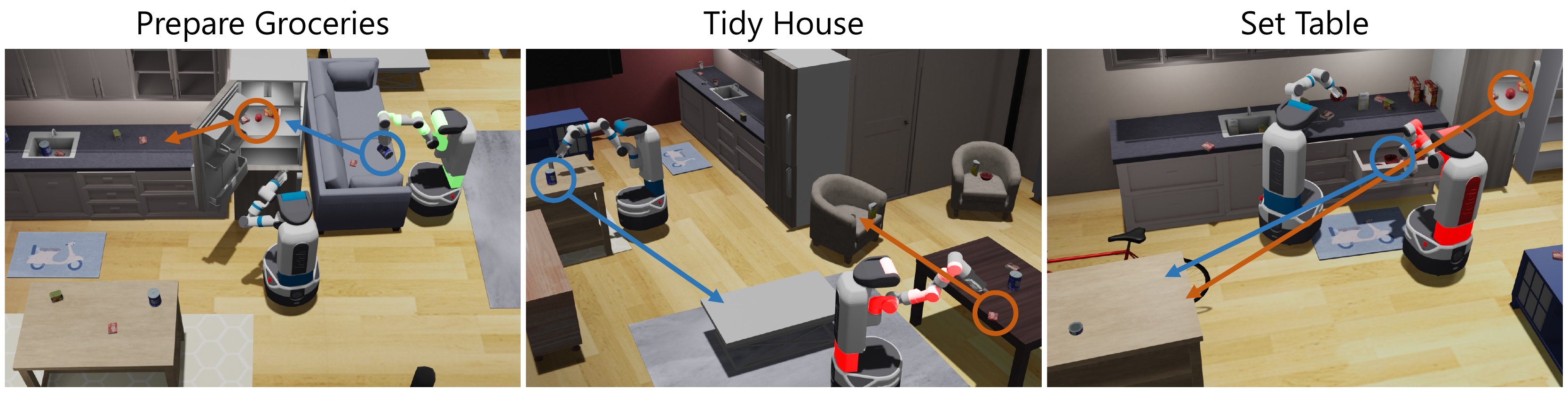}
  \caption{
    Overview of the tasks from \tname. 
    Circles indicate a possible starting location for each object and the arrows indicate the goal positions of where the objects should be moved to.
    Each task requires rearranging two objects.
    In Prepare Groceries, an object must be moved from the fridge to the counter and an object from a receptacle into the counter. In Tidy House, 2 objects starting on random receptacles throughout the house must be moved to a random goal receptacle.
    In Set Table, a bowl from the drawer and an apple from the fridge must be moved to the table. 
    Both the drawer and fridge start closed in Set Table. We refer to \cite{szot2021habitat} for detailed description and visualizations of the tasks.
    }
  \label{fig:task_overview}
\end{figure*}

\section{Additional \tname Details}
\label{sec:further_task_details} 
We follow the same setup as the original Home Assistant Benchmark~\cite{szot2021habitat}  but with modifications for multi-agent collaboration. We provide more details on \tname, particularly the episode setup, more details of the three tasks, and reward structure. See \Cref{fig:task_overview} for a visual overview of the tasks in \tname.

\textbf{Episode Setup:} At the start of each episode, both agents are randomly placed at a collision free location in the scene such that both agents start at least 2 meters apart.
The episode is successful if all target objects are placed within 15cm of their locations.
The episode fails if the agents collide or if the maximum episode horizon of 750 time steps is reached. 
We use the same ReplicaCAD scene split as prescribed by the Home Assistant Benchmark~\cite{szot2021habitat}  and the Habitat Rearrangement Challenge~\cite{habitatrearrangechallenge2022}.
For training in each task, we use 10,000 object configurations across 60 layouts of furniture in the scene.
While there are only 10,000 rearrangement problems, the agent is randomly spawned each episode, providing infinite variety. 
At evaluation time, we use 100 object configurations across 20 layouts of furniture in the scene, distinct from the furniture layouts in training.

\textbf{Set Table}: The goal of this task is to remove a bowl from a drawer, a fruit from the fridge, and  place the fruit in the bowl on the dinner table. 
Both the fridge and drawer are initially closed, and the robot must open them before removing objects. 
The fridge and the drawer are next to each other in the kitchen area of the scene.
The position of the dinner table relative to the kitchen changes depending on the scene.

\textbf{Tidy House:} The goal of this task is to move 2 objects from accessible initial locations to their target locations. 
The objects are spawned across 6 open receptacles, and assigned a goal from one of the 6 receptacles which is different from the starting receptacle. 
The receptacles start in random locations throughout the house and are always unobstructed to the agent accessing them.

\textbf{Prepare Groceries}: The goal of this task is to move 1 object from an open fridge to the counter and another object on the kitchen table to the fridge.
The counter and the fridge are always close to each other.
The vicinity of the table to the counter and fridge varies depending on the episode.

\textbf{Simulation}: We partially simulate physics to check for collisions between the robot and objects along with other robots. We kinematically move the robot base with a navigation mesh that defines the navigable regions of the scene according to static obstacles like furniture. We do not allow the robot to slide along obstacles. This setup has been shown to transfer well to the real world~\cite{Sim2RealHabitat}. The robots are able to collide with each other on the navigation mesh, which terminates the episode with failure.

\textbf{Reward Structure:}
The reward function for each task provides a sparse reward for task success, intermediate sparse rewards for completing subgoals, and a per time-step penalty.
This reward, described in \Cref{eq:reward}, is the same between all three \tname tasks.
\begin{align}
  \label{eq:reward}
  R(s_t) = 10 \cdot \mathbbm{1}_{\text{success}} + 0.5 \cdot  \mathbbm{1}_{\text{subgoal}} - 0.01 
\end{align} 
The first term, provides a $ +10$ reward for overall task success. 
We also provide a $ +0.5$ reward for either of the agents completing any sub-goal necessary for overall task success. 
These subgoals include picking up the target object, placing the target object at its goal, and for opening receptacles to access the object if it is necessary for the task as in \settable.
We also add a per time step negative penalty of $ -0.01$ to encourage more efficient solutions.
The reward is shared between both agents for a cooperative task.

\section{Comparing to Trajedi's Diversity Objective}
\label{sec:further_cmp} 
Here we highlight the differences between the diversity objective from the prior work of Trajedi~\cite{lupu2021trajectory} and the proposed \met. 
We assume both methods are latent variable conditioned policies with a discrete $ N$ dimension latent space with uniform prior $ p(z)$.
Let $ \pi(\tau | z) = \prod_{t=0}^{T} \pi(a_t | \tau_t, z)$ be the joint action probabilities for policy $ \pi$. 
The undiscounted Trajedi objective can be expressed as:
\begin{align*}
  \text{Diversity}^{\text{Trajedi}}(\pi) &= \text{JSD}(\pi(\tau | z_1), \dots, \pi(\tau | z_N)) \\
                                         &\propto \mathcal{H}\left( \sum_{z} \pi (\tau | z) \right)  - \sum_{z} \mathcal{H}\left( \pi(\tau | z) \right) 
\end{align*} 
Where the first term encourages the collective population to cover a diverse joint action distribution and the second term drives the individual policy joint action distribution to be as compact as possible.
Meanwhile the BDP objective is:
\begin{align*}
  \text{Diversity}^{\text{BDP}}(\pi) = -\mathcal{H}(p(z | s)) - \frac{1}{N} \sum_{z} \mathcal{H}(\pi(a | o, z))
\end{align*} 
The first term encourages the policy ID to be predictable from the state distributions generated by all policies.
In the second term, BDP encourages the opposite of Trajedi for each policy to be diverse.
This is because BDP does not need to balance a diversity of the overall population, which one policy could dominate.

In summary, Trajedi encourages diversity from the tension between the 2 objective terms, the first encouraging coverage of the trajectory space, while the second minimizes overlap between the policies.
BDP encourages diversity over observed behaviors while Trajedi encourages diversity over joint action distributions.

\section{\met Implementation Details}
\label{sec:method_details} 
In this section we provide further details about hierarchical policy training, policy architecture, and the \met discriminator architecture.

\subsection{\met Pseudocode}
\label{sec:pseudocode} 

\Cref{algo:bdp} presents the pseudocode for stage 1 and stage 2 training of \met.
Lines 1-5 initialize the behavior policy, discriminator, discriminator data buffer, and hyperparamters for training. 
Lines 7-11 train the behavior policy $ \pi^{b}$ by randomly sampling latents and then pairing the behavior policy with itself conditioned on these latents.
Updating $ \pi^{b}$ requires computing both the task reward and the diversity 
Next, lines 12-14 random sample trajectories from the discriminator data buffer and use them to update the discriminator to better predict the latent.
Then lines 17-19 update the \cag against the fixed $ \pi^{b}$.
Finally, the \cag is evaluated in ZSC.

\begin{algorithm}[H]
\begin{algorithmic}[1]
\STATE{Initial behavior policy $ \pi^{b}$}
\STATE{Initial discriminator network $ q_\phi$}
\STATE{Diversity objective weight $ \alpha$}
\STATE{Discriminator data buffer $ \mathcal{B}$ with 100k max size}
\STATE{Behavior latent prior $ p(z)$}
\FOR{each epoch in $ \pi^{b}$ training} 
  \STATE{$ z^1, z^2 \sim p(z)$}
  \STATE{Rollout out agent pair: $ (\pi_{z^1}^{b}, \pi_{z^2}^{b})$ }
  \STATE{Compute $ \mathcal{L}_J = J(\pi_{z^1}^{b}, \pi_{z^2}^{b}) $}
  \STATE{Compute  $ \mathcal{L}_D = \text{Diversity}(\pi^{b}) = \mathbb{E} \log q_\phi(z | \tau) + \mathcal{H}(a | o, z)$ }
  \STATE{Update $ \pi^{b} $ with PPO on objective $ \mathcal{L}_D + \alpha \mathcal{L}_D$}
  \STATE{Add $ (\tau^{1}, z^{1}), \left( \tau^{2}, z^{2} \right) $ to $ \mathcal{B}$}
  \STATE{Update $ \phi $ with random batches from $ \mathcal{B}$}
\ENDFOR

\STATE{Initial coordination policy $ \pi^{c}$}

\FOR{each epoch in $ \pi^{c}$ training} 
  \STATE{$ z \sim p(z)$}
  \STATE{Rollout out agent pair $ (\pi^{c}, \pi_{z}^{b})$ }
  \STATE{Update $ \pi^{c}$ with PPO on objective on $ J(\pi^{c}, \pi_{z}^{b}) $}
\ENDFOR
\STATE{Evaluate $ \pi^{c} $ in zero-shot coordination.}
\end{algorithmic}
\caption{\method pseudocode}
\label{algo:bdp}
\end{algorithm}

\subsection{Hierarchical Training}
In this work, all methods learn a high-level policy that control low-level skills. 
The high-level policy selects a discrete skill and a discrete parameterization for that skill.
All the possible skills are: open, pick, place, and navigate.
Each skill is parameterized by which entity to apply the action to.
In \tname, the possible entities are the target objects (2 for all tasks), the goal positions (2 for all tasks), and all possible receptacles in the scene (10 in total).
All tasks have the same action space.
We compute all possible actions given the compatibility of the action with the entity.
For example, the agent can never pick up the fridge so that is not a possible action. 
This gives a 21 element discrete high-level actions for each task and an additional 4 primitive actions for no-op, move-forward, turn-left, and turn-right.
At each step the agent selects from each of the 25 possible actions.

Training this high-level policy with on-policy RL requires changes to the RL training pipeline due to the separation between high-level actions and low-level robot actions.
The robot is executing low-level actions at every time step by controlling the base, arm, and gripper, but the high-level policy is not making decisions at every time step.
We only want to learn from the transitions where the high-level policy is acting in the environment.
Therefore, we change the rollout collection in PPO to conditionally write the transition to the rollout data storage if the high-level policy acted in that time step.
For example, when the robot is navigating, it is only executing base actions and the high-level policy is not acting 
These variable rollout sizes allow leveraging an efficient vectorized environment rollout implementation.

\subsection{Policy and Discriminator Network Architectures}
We first describe the policy neural network architecture.
The ResNet18~\cite{he2016deep} first encodes the $ 256\times 256$ depth visual observation to a 512-dimension vector. 
Next, these visual inputs are concatenated with an 18-dimension state information vector which includes the joint angles (8D) along with heading and distance to the object starts (4D), goals (4D), and other agent (2D).
These are then fed into a 2-layer LSTM~\cite{hochreiter1997long} with 512 hidden units.
This then produces a 512-dimensional vector which is separately fed into separate policy and value function networks. 
Each of these networks are a single layer linear layer.

All policies are trained with PPO~\cite{schulman2017proximal}.
DD-PPO~\cite{wijmans2019dd} is used to distribute training to 4 GPUs.
Each GPU runs 32 parallel environments and collects 128 simulation steps.
We run on NVIDIA V100 GPUs. 
We train the \bg in for 100 million low-level steps and the \cagent for another 100 million steps. 
In the second stage, we initialize the \cag, including the visual encoder, from scratch.

The discriminator is modeled as an MLP with 2 hidden layers with 512 hidden units. 
For the privileged state trajectories, the discriminator takes as input the robot $ x,y$ positions and a list of actions executed in a window spanning the last 40 steps.
In practice, it is difficult to tell the difference between different behavior latents from the first few time steps. 
Therefore, we do not provide the discriminator reward for the first $ 10\%$ of maximum time steps in the trajectory. 
We sample a new $ z^{1},z^{2}$
pair once every 10 policy updates for better training stability in both stages.
The discriminator is updated every policy update step with a buffer of at most the last 100,000 agent samples.

\subsection{Hyperparameter Selections}
\label{sec:hyperparam_details} 
For all methods we use the same hyperparameters unless stated otherwise.
For PPO policy optimization parameters, we use a learning rate of $ 0.0003$, $ 2$ epochs per-update, $ 2 $ mini-batches per-epoch, clip parameter of $ 0.2$, an entropy coefficient of $ 0.001$, and clip the gradient norm to a max value of $ 0.2$. 
For return estimation, we use a discount factor of $ \gamma=0.99$, GAE with $ \lambda=0.95$.
For the discriminator in \met we also use a learning rate of $ 0.0003$.
\met weighs the diversity reward by $ 0.01$ before adding it to the task reward.

\section{Additional Evaluation Details}
\label{sec:further_exp_details} 
We include supplementary information about how evaluation is conducted for \tname. Particularly, task plans for scripted holdout agents, how we implemented baselines, what we gave as input to \pbts, and how we obtained qualitative results.

\subsection{Scripted Holdout Agents}
\label{sec:further_zsc_details} 
Here we detail the task plans that the scripted agents execute for each task.
For every task, we include a scripted agent that only executes no-ops and two agents that execute a fixed portion of the task.
These two scripted agents will do half of the task involving interactions with only 1 object.
For example, in the \settable task, we have 1 scripted agent that will only rearrange the fruit, and another scripted agent that will only rearrange the bowl.
For training the learned coordination agents, we train them in two agent populations in the same manner as \xporacle.

\subsection{Baselines}
\label{sec:baseline_details} 
We include the necessary details for implementing baselines. 

\textbf{Self-Play (\splay)}~\cite{heinrich2016deep} and \textbf{Population-Based Training (\xp)}~\cite{jaderberg2017population} only vary in how they pair agents while training the population in stage 1. \splay only pairs the policies with themselves while \xp pairs policies with other policies in the population.
For \textbf{Fictitious Co-Play (\fcp)} we use 3 checkpoints from each agent population from the start, middle and end of training to the final learned population from stage 1, as in ~\cite{strouse2021collaborating}.
We then pair the \cag against these older agents in stage 2 training. 
With \textbf{Trajectory Diversity (\tdi) } \cite{lupu2021trajectory} we do not use any discounting factor in the JSD objective.

We fix the policy size to be the same between the behavior latent conditioned policy in \met and \emph{each} policy in the population of the baselines which maintain a set of $ N$ distinct policies.
While this increases the parameter count, we sufficiently train all policies to convergence in both stages with 100M steps of training experience per stage.

The hyperparameters are described in \Cref{sec:hyperparam_details}.

\subsection{Ground Truth State for \pbts Baseline}
\label{sec:gt_state} 
Here we describe the priviliedged ground truth state that the baseline \textit{\pbts (\Oracle, No-Vision)} takes as input. This ground truth state input consists of a set of binary predicates including: 
\begin{itemize}[itemsep=0pt,topsep=0pt,parsep=0pt,partopsep=0pt,parsep=0pt,leftmargin=*]
  \item $ \mathtt{robot\_at}(R, X)$ if robot $ R$ is at receptacle, object, or goal $ X$.
  \item $ \mathtt{is\_holding}(R)$ if robot $ R$ is holding an object.
  \item $ \mathtt{object\_at}(X, Y)$ if the object $ X$ is at goal location or receptacle $ Y$.
\end{itemize}
The truth values are enumerated for all possible inputs. This forms a 1D vector which is passed to the policy.
Both agents share the same fully observable state as input.

\subsection{Qualitative Result Details (Supplements \Cref{fig:zsc_study})}
\label{sec:qual_result_details} 
Here we include details of how we created \Cref{fig:zsc_study}, particularly, how we measured the portion of subgoals completed by the \cag in \Cref{fig:zsc_study}.
Let $ E_i$ denote the Bernoulli random variable that represents if the \cagent executed event $i$ when paired with a partner $\pi$. 
$\pi$ can belong to both the training population or the holdout population set. 
We then record $ p(E_i = 1 | \pi)$ to analyze how likely the \cag is to perform certain interaction types when paired with $\pi$, averaged over a 100 episodes. 
An agent biased towards an event $E_j$ would have a high $ p(E_j = 1| \pi)$ for all partner agents. 
An adaptive agent on the other hand, will have different $ p(E_j = 1| \pi)$, depending on its partner. 
Darker cells in \Cref{fig:zsc_study} indicate a higher value for $ p(E_1 = 1 | \pi)$, meaning the \cag is more likely to achieve this subgoal.

\section{Additional Experiments \& Ablations}
\label{sec:further_exps} 

\subsection{Extended ZSC Results (Supplements \Cref{tab:zsc})}
\label{sec:extended_zsc} 
In this section we present a more detailed breakdown of the ZSC results from \Cref{tab:zsc}.
Specifically, we separately show the ZSC performance between the scripted and learned unseen agents in the ZSC evaluation in \Cref{tab:zsc_all}.
These results indicate that in general ZSC is harder with scripted vs. learned unseen agents.
For example, in the \tdyhouse task, \met achieves $ 74.14\%$ success rate when paired with learned agents, but only $ 46.90\%$ success rate when paired with scripted unseen agents.
This same also trend holds for all other methods and tasks.

We experimented with the impact of different state inputs on \met's performance. We compared a version of \met, which takes RGB instead of depth images as input. We found that \met's performance remains mostly unchanged on the \prepgroc task when using RGB instead of depth images. The ZSC success rate of \met is $ 70 \%$ with RGB (averaged across 2 seeds) versus $ 76 \%$ with depth. We also evaluated an oracle version of \met, like \pbts, which takes the same privileged state information as input. On \prepgroc, this method achieves an average ZSC success rate of $ 80 \%$ (averaged across 2 seeds), compared to \pbts with $ 77 \%$ and non-privileged \met with $ 75 \% $ success rate.

\subsection{Policy Architecture Ablation (Supplements \Cref{tab:abl_discrim})}
\label{sec:policy_arch_abl} 
In \Cref{tab:abl_discrim}, we ablate the shared policy architecture in \met. 
The shared policy architecture lets \met be more sample-efficient by sharing weights, while generating behaviors through a behavior latent space $z$. We create two versions of \met: \\
\textbf{\metxdse} replaces behavior latent space $z$ with separate policies per agent, initialized randomly, but shares the visual encoder weights between all agents, to enable sample-efficiency. Essentially, the policies share the ResNets, but learn separate LSTM and MLP weights. \\
\textbf{\metxd} replaces the shared policy architecture with entirely different networks per agent initialized randomly, with no latent space, and no shared visual encoder. 
\metxd is the same as \pbt, but trained with a discriminator diversity reward from \met.
\begin{table}[h!] 
  \centering
  \resizebox{\columnwidth}{!}{
    \begin{tabular}{cccb}
      \toprule
      & \textbf{\metxdse} & \textbf{\metxd} & \textbf{\met} \\
\midrule
\traintitle &  50.09 {\scriptsize}$ \pm$ 0.01   &  39.19 {\scriptsize}$ \pm$ 0.00  &  74.81 {\scriptsize}$ \pm$ 0.01  \\
\evaltitle   &  33.40 {\scriptsize}$ \pm$ 0.06   &  26.76 {\scriptsize}$ \pm$ 0.06  &  46.43 {\scriptsize}$ \pm$ 0.08   
 \\
      \bottomrule
    \end{tabular}
  }
  \caption{
    Ablations on the shared policy architecture in \met.
  }
  \label{tab:abl_policy} 
\end{table}

\Cref{tab:abl_policy} shows that both the training and ZSC evaluation performance decrease as we decrease weight sharing through the behavior latent space, implying that both are essential for learning an adaptive \cagent.

\begin{table*}[ht] 
  \centering
  \resizebox{\textwidth}{!}{
    \begin{tabular}{raaccccc}
      \toprule
       & \textbf{\pbts} & \textbf{\xporacle} & \textbf{\splay} & \textbf{\xp} &  \textbf{\fcp} & \textbf{\tdi} & \textbf{\met} \\
      \midrule
      \multicolumn{7}{@{}l}{\textbf{Set Table}}\\
      \traintitle &  70.74 {\scriptsize$ \pm$ 0.00 }  &  90.52 {\scriptsize$ \pm$ 0.05 }  &  57.74 {\scriptsize$ \pm$ 0.01 }  &  46.67 {\scriptsize$ \pm$ 0.02 }  &  29.90 {\scriptsize$ \pm$ 0.07 }  & 43.24 {\scriptsize $ \pm$ 0.09} &  74.81 {\scriptsize$ \pm$ 0.01 }  \\
\evaltitle &  50.39 {\scriptsize$ \pm$ 0.09 }  & - &  17.97 {\scriptsize$ \pm$ 0.04 }  &  30.34 {\scriptsize$ \pm$ 0.04 }  &  37.50 {\scriptsize$ \pm$ 0.04 }  & 32.52 {\scriptsize $ \pm$ 0.04} &  \textbf{46.43 {\scriptsize$ \pm$ 0.08} }  \\
\HcUnseen &  35.94 {\scriptsize$ \pm$ 26.56 }  & - &  7.29 {\scriptsize$ \pm$ 5.02 }  &  27.08 {\scriptsize$ \pm$ 7.47 }  &  32.81 {\scriptsize$ \pm$ 1.56 }  & 26.17 {\scriptsize $ \pm $ 0.05} &  37.50 {\scriptsize$ \pm 2.31$ }   \\
\NnUnseen &  55.21 {\scriptsize$ \pm$ 8.49 }  & - &  21.53 {\scriptsize$ \pm$ 5.24 }  &  31.42 {\scriptsize$ \pm$ 4.95 }  &  38.28 {\scriptsize$ \pm$ 5.09 }  & 34.64 {\scriptsize $ \pm$ 5.17} &  47.92 {\scriptsize$ \pm$ 9.64 }  \\
 \\
      \midrule
      \multicolumn{7}{@{}l}{\textbf{Tidy House}}\\
      \traintitle &  74.90 {\scriptsize $ \pm$ 21.59  } &  92.28 {\scriptsize $ \pm$ 1.66  } &  34.18 {\scriptsize $ \pm$ 6.05  } &  36.13 {\scriptsize $ \pm$ 0.98  } &  12.04 {\scriptsize $ \pm$ 2.28  } &  39.65 {\scriptsize $ \pm$ 0.59  } &  73.83 {\scriptsize $ \pm$ 7.03  } \\
\evaltitle &  68.08 {\scriptsize $ \pm$ 0.09  } & - &  52.67 {\scriptsize $ \pm$ 0.06  } &  56.88 {\scriptsize $ \pm$ 0.07  } &  34.07 {\scriptsize $ \pm$ 0.09  } &  63.58 {\scriptsize $ \pm$ 0.05  } & \textbf{  66.71 {\scriptsize $ \pm$ 0.05  } } \\
\HcUnseen &  43.26 {\scriptsize $ \pm$ 15.71  } & - &  35.72 {\scriptsize $ \pm$ 9.14  } &  38.54 {\scriptsize $ \pm$ 11.52  } &  17.35 {\scriptsize $ \pm$ 11.29  } &  39.95 {\scriptsize $ \pm$ 8.87  } &  46.90 {\scriptsize $ \pm$ 9.11  } \\
\NnUnseen &  77.39 {\scriptsize $ \pm$ 10.11  } & - &  59.02 {\scriptsize $ \pm$ 7.69  } &  63.76 {\scriptsize $ \pm$ 8.75  } &  41.04 {\scriptsize $ \pm$ 10.80  } &  72.44 {\scriptsize $ \pm$ 5.08  } &  74.14 {\scriptsize $ \pm$ 3.61  } \\
 \\
      \midrule
      \multicolumn{7}{@{}l}{\textbf{Prepare Groceries}}\\
      \traintitle &  85.74 {\scriptsize $ \pm$ 2.82  } &  93.63 {\scriptsize $ \pm$ 0.28  } &  47.07 {\scriptsize $ \pm$ 29.88  } &  69.34 {\scriptsize $ \pm$ 1.76  } &  44.40 {\scriptsize $ \pm$ 6.38  } &  34.56 {\scriptsize $ \pm$ 27.94  } &  89.67 {\scriptsize $ \pm$ 2.51  } \\
\evaltitle &  77.01 {\scriptsize $ \pm$ 0.05  } & - &  56.04 {\scriptsize $ \pm$ 0.07  } &  56.08 {\scriptsize $ \pm$ 0.09  } &  30.00 {\scriptsize $ \pm$ 0.07  } &  53.84 {\scriptsize $ \pm$ 0.08  } & \textbf{  75.85 {\scriptsize $ \pm$ 0.05  } } \\
\HcUnseen &  60.00 {\scriptsize $ \pm$ 0.00  } & - &  54.86 {\scriptsize $ \pm$ 17.89  } &  50.70 {\scriptsize $ \pm$ 15.56  } &  29.06 {\scriptsize $ \pm$ 12.63  } &  42.09 {\scriptsize $ \pm$ 17.53  } &  60.42 {\scriptsize $ \pm$ 15.76  } \\
\NnUnseen &  83.38 {\scriptsize $ \pm$ 6.45  } & - &  56.49 {\scriptsize $ \pm$ 7.18  } &  58.10 {\scriptsize $ \pm$ 11.37  } &  30.71 {\scriptsize $ \pm$ 7.31  } &  58.25 {\scriptsize $ \pm$ 9.74  } &  81.64 {\scriptsize $ \pm$ 3.52  } \\
 \\
      \bottomrule
    \end{tabular}
  }
  \caption{
    Detailed breakdown of the ZSC success rates from \Cref{tab:zsc} by unseen agent type: scripted or learned.
    Across most methods and tasks, methods achieve lower success rates when paired with unseen scripted agents.
  }
  \label{tab:zsc_all} 
\end{table*}

\end{document}